\documentclass[final]{cvpr}
\usepackage{times}
\usepackage{epsfig}
\usepackage{graphicx}
\usepackage{amsmath}
\usepackage{amssymb}
\usepackage{animate}
\usepackage{caption}
\usepackage{adjustbox}

\newcommand{\bw}{{\mathbf{w}}}

\newcommand{\by}{{\mathbf{y}}}

\newcommand{\calL}{{\mathcal{L}}}

\newcommand{\btheta}{\mbox{\boldmath $\theta$}}

\newcommand{\be}{\begin{eqnarray}}
\newcommand{\ee}{\end{eqnarray}}
\newcommand{\bee}{\begin{eqnarray*}}
\newcommand{\eee}{\end{eqnarray*}}

\newcommand{\matrixb}{\left[ \begin{array}}
\newcommand{\matrixe}{\end{array} \right]}   

\newcommand{\Tref}[1]{Table~\ref{#1}}
\newcommand{\Eref}[1]{Eq.~(\ref{#1})}
\newcommand{\Fref}[1]{Fig.~\ref{#1}}

\newcommand{\Cref}[1]{Chap.~\ref{#1}}

\def\eg{\emph{e.g. }}

\def\etal{\emph{et al. }}
\def\ie{\emph{i.e. }}

\usepackage{booktabs}
\usepackage{mathtools}
\usepackage{pifont}

\usepackage[table,x11names]{xcolor}

\usepackage[breaklinks=true,colorlinks,bookmarks=false]{hyperref}

\begin{document}

\title{Restoration of Video Frames from a Single Blurred Image with Motion Understanding}


\author{%
	Dawit Mureja Argaw, Junsik Kim, Francois Rameau, Chaoning Zhang, In So Kweon \\
    KAIST Robotics and Computer Vision Lab., Daejeon, Korea \\
}
\maketitle
\thispagestyle{empty}

\begin{abstract}
We propose a novel framework to generate clean video frames from a single motion-blurred image.
While a broad range of literature focuses on recovering a single image from a blurred image, in this work, we tackle a more challenging task \ie video restoration from a blurred image.
We formulate video restoration from a single blurred image as an inverse problem by setting clean image sequence and their respective motion as latent factors, and the blurred image as an observation.
Our framework is based on an encoder-decoder structure with spatial transformer network modules to restore a video sequence and its underlying motion in an end-to-end manner.
We design a loss function and regularizers with complementary properties to stabilize the training and analyze variant models of the proposed network.
The effectiveness and transferability of our network are highlighted through a large set of experiments on two different types of datasets: camera rotation blurs generated from panorama scenes and dynamic motion blurs in high speed videos.
\end{abstract}

\vspace{-2mm}
\section{Introduction}
\vspace{-1.5mm}
Capturing an image is not an instant process; to capture enough photons, the photosensitive elements of a camera have to be exposed to light for a certain interval of time, called exposure time. Therefore, during this interval if an object is moving in the observed scene or the camera is undergoing an arbitrary motion, the resulting image will contain a blurring artifact known as \emph{motion blur}. In general, motion blur is an unwanted behaviour in vision applications \eg image editing \cite{gunturk2012image}, visual SLAM \cite{lee2011simultaneous} and 3D reconstruction \cite{seok2013dense}, as it degrades the visual quality of images. To cope with this type of artifact, image deblurring aims to restore a sharp image from a blurred image. This problem is known to be ill-posed since the blur kernel used for deconvolution is generally assumed to be unknown. 

Earlier studies assume a uniform-blur over the image to simplify the estimation of the single deconvolution blur kernel used to remove the blur \cite{fergus2006removing,cho2009fast,levin2009understanding}. Even though the methods deploy deblurring tasks with uniform-blur assumption, the assumption is often violated in practice. For instance, when the blur is caused by out-of-plane camera rotation, the blur pattern becomes spatially variant. Moreover, the problem is more complex when objects in a scene are moving \ie dynamic blur. While previous literature focuses on recovering a sharp image from a blurred image, we tackle a more challenging task \ie video restoration from a blurred image.

Restoring the underlying image sequence of a blurred image requires both contents and motion prediction. We formulate video restoration from a blurred image as an inverse problem where a clean sequence of images and their motion as latent factors, and a blurred image as an observation. Some of previous deblurring approaches \cite{Kim_2014_CVPR,Zhang_2015_CVPR,sellentECCv2016,wenqi17iccv,park17iccv,argaw2021optical} also estimate the underlying motion in a blurred image, however, their goal remains in single frame restoration. Recently Jin \etal \cite{Jin_2018_CVPR} proposed to extract video frames from a single motion-blurred image. Their approach is close to image translation model without inferring underlying motions between the latent frames. Purohit \etal \cite{purohit2019bringing} addressed this issue by estimating pixel level motion from a given blurred input. However, their model is still prone to sequential error propagation as frames are predicted in a sequential manner using a deblurred middle frame. Our work differs from previous works in two aspects. First, we use a single network to restore the underlying video frames from a single motion-blurred image in an end-to-end manner while \cite{Jin_2018_CVPR,purohit2019bringing} jointly optimize multiple networks for the task. Second, our approach is not explicitly dependent on a deblurred middle frame in order to restore non-middle frames, and hence, is relatively robust to sequential error propagation which occurs due to erroneous middle frame.

In this paper, we propose a novel framework to generate a clean sequence of images from a single motion-blurred image.
Our framework is based on a single encoder-decoder structure with Spatial Transformer Network modules (STN) and Local Warping layers (LW) to restore an image sequence and its underlying motion. Specifically, a single encoder is used to extract intermediate features which are passed to multiple decoders with predicted motion from STN and LW modules to generate a sequence of deblurred images. We evaluate our model on two types of motion blur. For rotation blur, which is caused by abrupt camera motion, we generated a synthetic dataset from panoramic images \cite{jxiaoCVPR2012}. For dynamic blur caused by fast moving objects in a scene, we used a high speed video dataset \cite{Nah_2017_CVPR}. The proposed model is evaluated on the panorama and the high speed video datasets under various motion patterns. Both the quantitative metrics and qualitative results highlight that our method is more robust and performs favorably against the competing approaches~\cite{Jin_2018_CVPR} We also provide comparison with single image deblurring approaches on GoPro benchmark dataset \cite{Nah_2017_CVPR} to evaluate the performance of the middle frame prediction. For further investigation, we demonstrate the transferability of our model by cross-dataset evaluation. 

In short, our contributions are as follows.
1) We propose a novel unified architecture to restore clean video frames from a single motion-blurred image in an end-to-end manner.
2) A simple yet effective mechanism is presented to generate a realistic rotational blur dataset from panoramic images
3) We carefully design loss terms for stable network training and perform thorough experiments to analyze the transferability and flexibility of the proposed architecture.
4) Our model quantitatively and qualitatively performs favorably against the competing approaches.

\vspace{-2mm}
\section{Related Works}
\vspace{-2mm}
\label{sec:related}
\paragraph{Image deblurring.}
Image deblurring is an ill-posed inverse problem when a blur kernel is unknown \ie blind deconvolution problem, as different latent images can be transformed to a blurred image depending on its blur kernel. Early stage of deblurring studies \cite{cho2009fast,fergus2006removing,pan2014deblurring,michaeli2014blind,pan2016blind,chakrabarti2016neural,dong2017blind,yan2017image} assume a single blur kernel that is applied to an image globally. The restoration of blur images is often modeled as a maximization problem of probabilistic models~\cite{cho2009fast,fergus2006removing}. To narrow down the ambiguity of the blur kernel estimation, natural image priors~\cite{michaeli2014blind,pan2014deblurring,pan2016blind,yan2017image} are exploited. While single blur kernel estimation approaches are effective when blur kernels are shift-invariant, they fail when the blur is not spatially uniform. To restore images affected by motion blur from pure rotations, Dong \etal \cite{dong2017blind} use the geometric information of the camera motion as a prior to recover the non-uniform blur model. Recently, deep network based methods \cite{Nah_2017_CVPR,Zhang_2018_CVPR} are proposed to handle general blur patterns without the uniform blur assumption. 
Nah \etal  propose multi-scale deep networks with multi-scale loss that mimics coarse-to-fine approaches to restore sharp images under non-uniform blurred images. Zhang \etal proposed a spatially variant neural networks to learn spatially variant kernels. However, the approaches addressed here only recover a single image while our goal is to recover the underlying sequence of frames from a given blurred image.

\vspace{-3mm}
\paragraph{Sequence restoration from a blurred image.}

Recently, Jin \etal\cite{Jin_2018_CVPR} proposed to extract a video sequence from a single motion-blurred image using multiple deep networks. They showed that deep networks can successfully generate an image sequence from a blurred image, however there remains a few limitations. Their proposed framework consists of multiple networks of which each network is specialized to predict a specific frame in a sequence. Each network is trained separately and sequentially starting from the middle frame and then adjacent frames taking previously predicted frames as inputs. As a result, the non-middle frame prediction heavily relies on previously predicted frames including the middle frame itself, therefore when the middle frame is erroneous the error propagates across frames. Purohit \etal\cite{purohit2019bringing} proposed a two-step strategy to generate a video from a motion-blurred image using three complementary networks. They used video autoencoder to learn motion and frame generation from clean frames as a pretraining phase. Later, they introduced a motion disentangle network to extract motion from blurred image. They also used independent deblurring network as their approach requires a clean middle frame generated from a blurred image in advance. Although their approach takes motion information into account, the approach generates frames sequentially starting from the middle frame to adjacent frames which results in error propagation just as in \cite{Jin_2018_CVPR}. Unlike the previous works, our approach runs in an end-to-end manner within a single training stage without error propagation across frames.

\section{Dataset}
\label{sec:dataset}
\vspace{-1.5mm}
Collecting a large number of natural motion-blurred images is a daunting task. Hence, a common practice in computer vision research is to generate blurry images by combining a sequence of sharp images using various approaches ranging from simple averaging~\cite{Nah_2017_CVPR,Jin_2018_CVPR} to learnable methods~\cite{BrooksBarronCVPR2019}. The source of motion blur in an image can be generalized into two main categories: rapid camera motion (camera shake) and  dynamic motion of objects in the scene. In this section, we briefly explain how we generate a blurry image dataset by considering each case individually. 
\vspace{-2mm}
\begin{figure*}[!t]
\begin{center}
\setlength{\tabcolsep}{0.8pt}
\resizebox{1.0\linewidth}{!}{
\begin{tabular}{cccc}
         \includegraphics[width=0.17\linewidth]{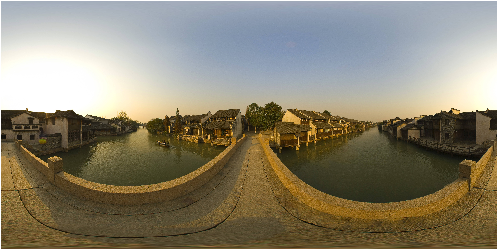} &
         \includegraphics[width=0.1\linewidth]{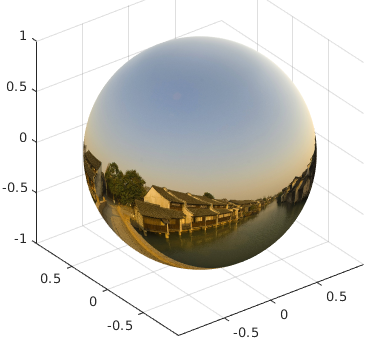} &
         \includegraphics[width=0.1\linewidth]{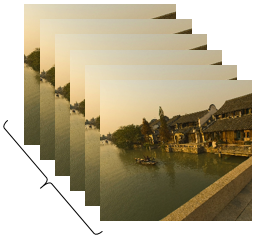} &
         \includegraphics[width=0.075\linewidth]{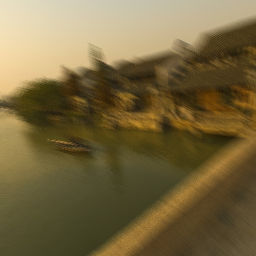} \vspace{-2mm} \\
         \tiny(a) & \tiny(b) & \tiny(c) & \tiny(d)
         \end{tabular}}
\end{center}
\vspace{-5mm}
\caption{\textbf{Rotational blur dataset generation}. (a) input panorama image, (b) panorama projection on a unit sphere, (c) intermediate frames between the initial and final images (d) blurred image obtained by averaging the captured frames.}
\label{fig:pano}
\vspace{-4mm}
\end{figure*}
\paragraph{Rotational blur (synthetic).}
In order to generate a rotation blurred image dataset, we use the SUN360 panorama dataset~\cite{jxiaoCVPR2012}. This dataset provides various panoramas with $360^{\circ}$ field of view. Hence, a virtual camera can be modeled to point at different orientations to represent the camera rotation in $SO(3)$. Given a panorama $P$ of size $H \times W$, we developed a simple yet effective framework to generate blurred images. First, the panorama is projected onto a unit sphere by linearly mapping each pixel coordinate $(x,y) \in P$ into spherical coordinates $(\theta,\phi)$ with $\theta\in(0,2\pi)$ and $\phi \in (-\pi/2, \pi/2)$. Then, a synthetic image can be captured via a virtual camera by re-projecting the 3D points on the sphere into an image plane as briefly discussed in~\cite{icraMeiR07} and~\cite{oleksandrcvmp}. Using this procedure we first capture an image by positioning the virtual camera at an arbitrary orientation. We call the image generated at this orientation \textit{initial image}. Then, we rotate the camera by a random rotation matrix (with $\beta = (\beta_x, \beta_y,\beta_z)$ its Euler angle representation) and capture a second image at the new camera position called \textit{final image}. We finally use a quaternion spherical linear interpolation technique (Slerp)~\cite{slerpref} to 

capture intermediate frames between the initial and final images. All the resulting images (initial, final and intermediate frames) are then averaged to generate a blurry image. The camera rotation angle is uniformly sampled from [$-10^{\circ}, 10^{\circ}$]. In order to generate a realistic blurred image, the number of intermediate images have to be adjusted automatically depending upon the rotation magnitude between the initial and final frames. Therefore, we use a simple linear relationship between the number of frames to be generated ($n$) and the rotation magnitude as follows: $n = c + \frac{1}{3} \| \mathbf{\beta} \|$, where $c$ is a constant and $\| \mathbf{\beta} \|$ is the magnitude of $\beta$. In this manner, we use $1000$ panoramic images from which we generate $26,000$ training and $3,200$ test images of size $128\times128$px. The dataset generation process is summarized in \Fref{fig:pano}. 
\vspace{-3mm}
\paragraph{Dynamic motion (real).}
In order to generate more realistic and generic (arbitrary camera motions and dynamic scene) blurred images, we take advantage of a GoPro high speed video dataset~\cite{Nah_2017_CVPR}. This dataset provides 22 training and 11 test scenes, each scene containing frames of size $1280\times720$px. A blurry image is generated by averaging $n$ consecutive frames~\cite{Nah_2017_CVPR,Jin_2018_CVPR}. In our experiments, we fixed $n=7$ and generated $20,000$ training images by randomly cropping images of size $256\times256$px. We also generated $2000$ test images from the test videos by averaging 7 consecutive frames.


\section{Method}
\label{sec:method}
\begin{figure*}[!t]
	\centering
	\includegraphics[width=0.9\textwidth,trim={12.2cm 7cm 16.5cm 9.0cm},clip]{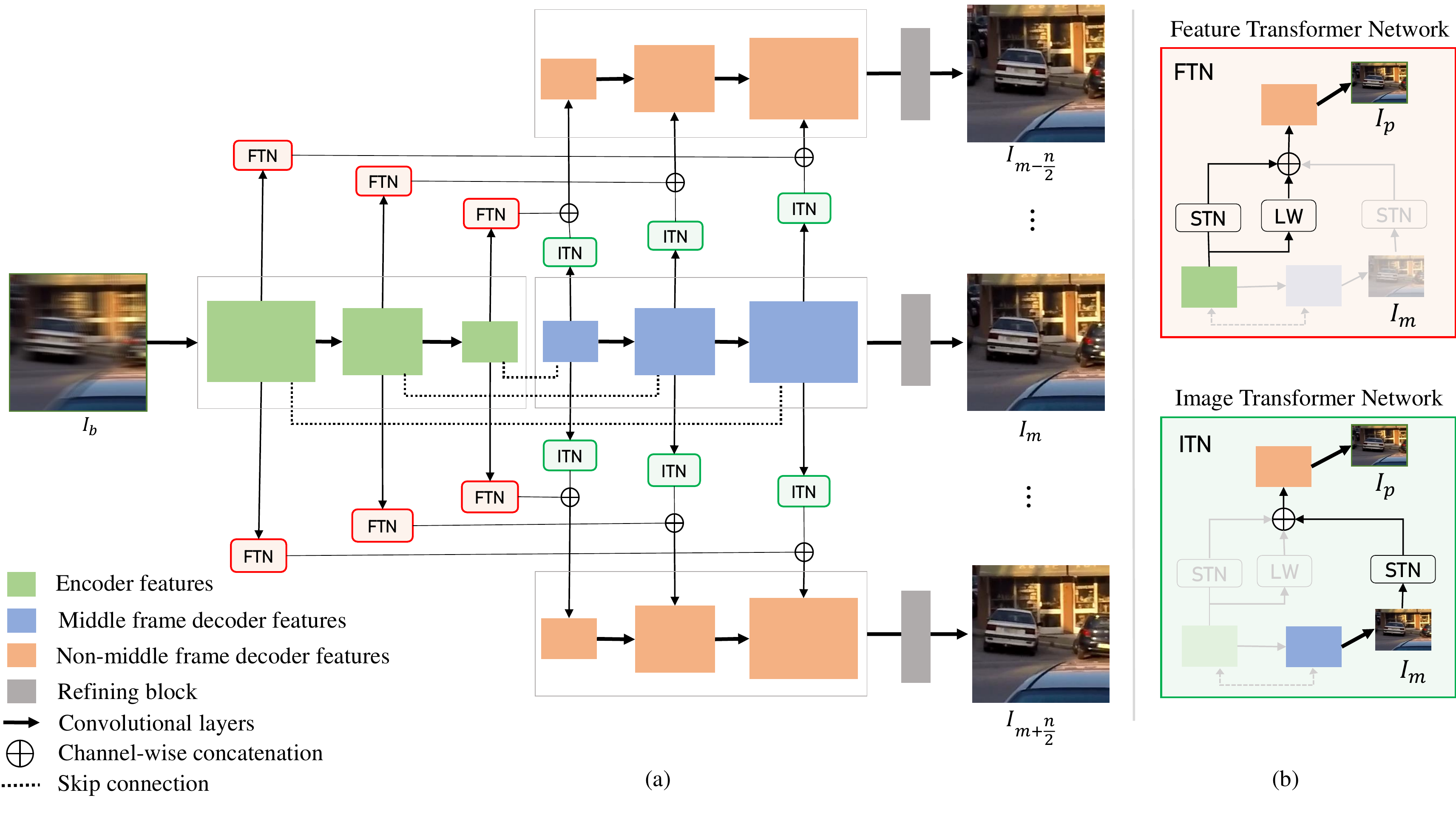}
\caption{\textbf{Overview of our network}. (a) The middle frame is predicted using an encoder-decoder structure. The non-middle frames are reconstructed by transforming the multi-layer features of the middle frame. (b) Feature transformer network (FTN) transforms features locally via local warping (LW) and globally via spatial transformer network (STN). Image transformer network (ITN) transforms predicted middle frame via STN. Finally, the predicted frames are passed through a refining network.}
	\label{fig:model}
	\vspace{-3mm}
\end{figure*}

\vspace{-2mm}
Given a blurry image $I_b$ synthesized from averaging $n$ latent frames, deblurring approaches predict the middle latent frame $I_m$. In this work, we restore the entire latent frame sequence 
$\{I_{m-\frac{n}{2}},\ldots,I_{m-1},I_m,I_{m+1},\ldots,I_{m+\frac{n}{2}}\}$, where $I_m$ is the deblurred middle frame and $\{I_{j}\}_{j=m-\frac{n}{2}}^{m+\frac{n}{2}}$ where $j \neq m$  are the recovered non-middle latent frames. The input blur is used as a motion cue to decode non-middle latent frame features (with respect to the middle latent frame) using transformer networks as shown in \Fref{fig:model}.

\subsection{Middle latent frame}
The middle latent frame $I_m$ is reconstructed using a U-net~\cite{RFB15a} like network. The \textit{encoder} contains five convolutional blocks, each block containing two layers of convolutions with spatial kernel size of $3\times3$ and stride size of 2 and 1, respectively. It outputs encoded features at different feature levels as shown in \Fref{fig:model}a. The encoded features are then decoded to predict the middle latent frame. The middle frame decoder network also contains five convolutional blocks to upsample features and to predict images at different scales. In each block, a feature is first upscaled using a deconvolution layer of kernel size $4\times4$ and a stride size of 2. The image predicted in the previous block is also upscaled in the same manner. The upsampled feature and its respective image are then concatenated channel-wise with the corresponding feature from the encoder (skip connection as shown in the \Fref{fig:model}a), then  passed through five layers of convolutions with dense connections to output a feature, which will be used to predict an image at current block. In this manner, features and images are successively upsampled to predict a full scale middle frame. Along with the last feature map from the decoder, the predicted image is finally passed through a \textit{refining} convolutional block. The purpose of this network is to further refine the predicted frame with contextual information by effectively enlarging the receptive field size of the network.

\subsection{Non-middle latent frame}
\vspace{-0.5mm}
The non-middle latent frames are reconstructed from the encoded features via learned transformations by feature transformer networks (FTN) and image transformer networks (ITN) as shown in \Fref{fig:model}b. 
\vspace{-2.5mm}
\paragraph{Feature transformer network.}
The feature transformer network inputs an encoded feature and transforms it into a non-middle latent feature in accordance with the learned motion. It consists of spatial transformer network (STN) \cite{jaderberg2015spatial}
and local warping (LW) layer. The STNs learn to estimate global transformation parameter $\theta_{[R|T]}$ from encoded features of a motion-blurred input and transform them accordingly. In order to compensate for locally varying motions, we designed a local warping network. This network is conditioned on the input feature like STN, however, instead of predicting global transformation parameters, it predicts pixel-wise displacement \ie \textit{motion flow}. Given an input feature $U\in \mathbb{R} ^{H\times W\times C}$, the local warping network outputs a motion flow of size $H\times W\times 2$. By warping the input feature with the predicted motion flow, we obtain a locally transformed feature which is concatenated with the globally transformed feature as shown in \Eref{eqn:decoder1}.
\begin{equation}
U^l_t = \mathrm{STN}^l (U_e^l) \oplus \mathrm{LW}^l (U_e^l), \qquad
\label{eqn:decoder1}
\end{equation}
where $l=\{1,\; ...\; ,k\}$ is an index for $k$ feature levels, $U_e$ is an encoded feature and $U_t$ is a transformed feature.
\vspace{-2mm}
\paragraph{Image transformer network.}
The middle frame decoder predicts frames at different feature levels. To guide the reconstruction of the non-middle latent frames with respect to the  middle frame, we used STNs to spatially transform the estimated middle frames according to the learned inter-frame motion \ie $I^l_t = \mathrm{STN}(I_m^l)$, where $I_m$ is the predcited middle frame and $I_t$ is the transformed image (see \Fref{fig:model}b). 

FTNs decode non-middle latent features from encoded features via learned local and non-local motions while ITNs globally capture the motion of the non-middle latent frame relative to the middle latent frame. The outputs of both networks are aggregated channel-wise and are passed through a decoder to predict a non-middle frame (\Fref{fig:model}b). We also input the encoded feature into the non-middle frame decoder in order to guide the decoder to learn the spatial relation between the middle and the non-middle frame as shown \Eref{eqn:decoder2}.
\vspace{-1mm}
\begin{equation}
I_p^l = \mathcal{D}^l(U_t^l \oplus I_t^l \oplus U_e^l)
\label{eqn:decoder2}
\end{equation}

where $p  = \{m-\frac{n}{2},\ldots,m-1,m+1,\ldots,m+\frac{n}{2}\}$ is an index for non-middle latent frames and $\mathcal{D}$ is a non-middle frame decoder.

Given ground truth non-middle frames during training, our model learns the transformation parameters to be applied to the encoded features of a blurry input at different scales in order to output the desired non-middle frames. The fact that unique transformer networks are applied at each feature and image scale gives the model a capacity to learn various types of transformations, hence, making it robust to different blur patterns including large blurs. 

\subsection{Loss functions}
\vspace{-0.5mm}
To ensure stable training and to restore clean latent frame sequences in a temporally coherent manner, we carefully designed the following loss functions,
\vspace{-2mm}
\paragraph{Photometric loss.}

For sharp video frame reconstruction, we trained our network with a weighted multi-scale photometric loss between the images predicted by the decoder network and the ground truth image. A bilinear downsampling is used to resize the ground truth image to the corresponding predicted frame size at different scales. Let $\{\hat{y}\}_{l=1}^k$ denote a set of predicted images from the smallest size ($\hat{y}_1$) to the full scale ($\hat{y}_k$), and $\{y\}_{l=1}^k$ represent a set of downsampled ground truth images where $y_k$ is a full scale ground truth image. For training a model predicting a sequence with $n$ frames from a single blurry image, we compute multi-scale photometric loss as follows,
\vspace{-2mm}
\begin{equation}
     \calL_{mp} = \sum_{j = 1}^{n}\sum_{l=1}^{k}{\bw_l \cdot\big|\by_{j,l} -\hat{\by}_{j,l}\big|_1}
     \label{eqn:pml}
     \vspace{-2mm}
\end{equation}
where $\bw_l$ is the loss weight coefficient for feature level $l$ and $j$ is an index for frame sequence.
 \vspace{-2mm}
\paragraph{Transformation consistency loss.}
We used individual transformer networks at each feature level when predicting non-middle frames. This augments our model with the capacity to learn transformations at different levels making it robust to various blur patterns. However, we expect the transformations at different scales to be aligned for successfully reconstructing temporally consistent non-middle frames. Especially at the initial stages of the training where the transformer parameters are random, it is beneficial that our model understands the relationship between the transformations across different frame levels. In order to impose this notion into our model and facilitate a smooth training, we propose the \textit{transformation consistency loss}. Let $\{\btheta\}_{l = 1}^k$ be the set of predicted transformation parameters at different scales. The transformation consistency loss for predicting $n-1$ non-middle frames can be defined as the term $\calL_{tc}$ in \Eref{eqn:tc},

where $|.|_2$ is an $\ell2$ loss between the transformation parameters.
\vspace{-4mm}
\begin{equation}
    \calL_{tc} = \sum_{j = 1}^{n-1}\sum_{l=2}^{k}{\big|\btheta_{j,l} - \btheta_{j,l-1}\big|_2}
    \label{eqn:tc}
     \vspace{-3mm}
\end{equation}
\paragraph{Penalty term.}
 Predicting multiple frames from a single blurry image can be problematic at times when the model fails to learn any type of transformation and simply replicates the middle frame predicition as non-middle frames. In order to remedy this issue, we design a penalty term to enforce symmetric diversity among generated images. This is accomplished by explicitly maximizing the sum of absolute difference (SAD) \ie~minimizing the negative SAD between a predicted frame and its time-symmetric (about the middle frame) ground truth frame. For example, when predicting seven frames $\{I_1,...,I_4,....,I_7\}$, we enforce the predicted image $I_1$ to be different content-wise from the ground truth image $I_7$ and vice versa. The penalty is imposed in a symmetric manner (as a matter of design choice inspired by the network architecture) such that the model learns to be sensitive to smaller transformations close to the middle frame as well as larger transformations at the end frames. Given a predicted frame $\hat y_i$ and the corresponding time-symmetric ground truth $y_{n+1-i}$, the penalty term is computed as the term $\calL_p$ in \Eref{eqn:pt},

where $m$ is the middle frame index, $n$ is the total number of frames. 
\vspace{-2mm}
\begin{equation}
    \calL_p = -\sum_{j = 1,j \neq m}^{n}\big|y_{n+1-j} - \hat y_j\big|_1
\label{eqn:pt}
\end{equation}
The final training loss function is defined as follows,
\begin{equation}
    \calL = \calL_{mp} + \lambda_{tc}\calL_{tc} + \lambda_p \calL_{p},
\end{equation}
where $\lambda_{tc}$ and $\lambda_{p}$ are weight coefficients for transformation consistency loss and penalty term, respectively.

\vspace{-1.5mm}
\paragraph{Temporal ambiguity and network training.}
The task at hand has two main ambiguities. i. \emph{temporal shuffling} and ii. \emph{reverse ordering}.
As explained in section 3, motion-blur is the result of an averaging process and, restoring temporally consistent (no shuffling) sharp frame sequence from a given motion-blurred input is a non-trivial task as the averaging destroys the temporal order. Jin \etal \cite{Jin_2018_CVPR}  mentions that  photometric loss is not a sufficient constraint to make their network converge. Hence, they propose a pair-wise order invariant loss to train their network. Purohit \etal \cite{purohit2019bringing} also uses the same loss function to fine-tune the recurrent video decoder in their network.

We experimentally find that a multi-scale photometric loss is a sufficient constraint to train our network. We further impose more constraints using other loss terms to improve performance (see Ablation studies). By design nature, our model allows motions to be learned in a symmetric manner (about the middle frame) with transformer networks close to the middle frame decoding smaller motions and those further from the middle frame decoding larger motions. This notion is enforced by symmetric constraint term and transformation consistency loss during training. The fact that our model is optimized in a joint manner allows frames to be reconstructed in a motion-guided sequence.

Other than \emph{temporal shuffling}, another issue is \emph{reverse ordering}. Given a single motion-blurred input, recovering ground truth order is a highly ill-posed problem which is intractable since reversely ordered frames result in the same motion-blurred image. Neither our work nor previous works \cite{Jin_2018_CVPR,purohit2019bringing} are capable of predicting the right order. Hence, we evaluate frame reconstructions using both ground truth order and its reverse order, then report the higher metric in the experiment section. A recent work by Argaw \etal~\cite{argaw2021motionblurred} proposed an optical flow based approach to reconstruct sharp frames in a temporally ordered manner, however, their approach requires at least two blurry frames.


\begin{table}[!t]
\setlength{\tabcolsep}{7pt}
\caption{Quantitative evaluation on Panorama blur dataset}
\label{tbl:quant1}
\centering
    \begin{tabular}{l|lccc}
    \toprule
    & Methods & $F_i$ & $F_m$ & $F_f$  \\ 
    \midrule
    PSNR & Jin \etal  &  22.007 & 22.493 & 22.157  \\
    & Ours & \textbf{23.693} & \textbf{24.049} & \textbf{23.874} \\ \midrule
    SSIM &  Jin \etal  &  0.572 & 0.621  & 0.589\\
    & Ours & \textbf{0.699}  & \textbf{0.716}& \textbf{0.704}\\
    \bottomrule
    \end{tabular}
    \vspace{-1.5mm}
\end{table}

\vspace{-1.5mm}
\section{Experiment}
\vspace{-1mm}
\label{sec:experiment}
\paragraph{Implementation details}
Our model is implemented using PyTorch \cite{paszke2017automatic}. We chose Adam \cite{KingmaB14} as an optimizer with $\beta_1$ and $\beta_2$ fixed to 0.9 and 0.999, respectively. On our synthetic blur dataset, we train the model using images of size $128\times128$px and a mini-batch size of 8 to predict initial, middle and final frames. A mini-batch size of 4 and input size of $256\times256$px is used to predict sequences of frames when training on the high speed video dataset. In all experiments, we train our model for 80 epochs. We set the learning rate $\lambda = 1e-4$ at the start of the training and decay it by half at epochs 40 and 60.
All the training images are cropped from the original resolution images without resizing.

\subsection{Video restoration results}
\vspace{-0.5mm}
In this section, we analyze the performance of our model for sequential frame restoration qualitatively and quantitatively on both camera shake blurs generated from panoramic scenes and dynamic blurs obtained from high speed videos.

\vspace{-2mm}
\paragraph{Quantitative evaluation.}
We report test results using peak signal-to-noise ratio (PSNR) and structural similarity (SSIM) metrics. To purely evaluate the quality of generated images without ordering estimation issue due to \emph{reverse ordering}, we report the higher PSNR/SSIM metric of either ground truth order or reverse order of frames \ie $\textrm{max}\{\textrm{PSNR/SSIM}(F_i \rightarrow F_f), \textrm{PSNR/SSIM}(F_f \rightarrow F_i)\}$, where $F_i$, $F_m$ and $F_f$ refer to the initial, middle and final frames in the restored sequence, respectively. We compared our approach with previous works \cite{Jin_2018_CVPR} on both rotational and dynamic blur datasets as tabulated in \Tref{tbl:quant1} and \Tref{tbl:quant2}. On Panorama blur dataset, our model outperforms Jin \etal by 1.65 dB on average. The middle and non-middle frame accuracy are similar on average (see \Tref{tbl:quant1}) mainly because rotational blurs are static blurs with uniform camera motion. Hence, it is relatively easier for the network to infer the global motion and decode
frames accordingly. In contrast, the GoPro blur dataset, however, contains arbitrary camera motions with dynamic scene and hence, decoding frames require inferring non-uniform global and local motions between frames (with middle frame as a reference). Therefore, the network reliably performs for middle frame prediction and performs less for the end frames due to randomness of motions (see \Tref{tbl:quant2}). On GoPro blur dataset, our model outperforms Jin \etal by 2.51 dB on middle frame prediction and by 3.69 dB on non-middle frame predictions. This highlights the advantage of adopting a motion-based approach to leverage blur as a motion cue to decode latent frames rather than extracting frames sequentially in a generic manner. 

The performance gap between the middle frame and non-middle frames is relatively larger in Jin \etal than our method. This is due to sequential prediction in Jin \etal which makes non-middle frame prediction heavily dependent on the generated middle frame, resulting in error propagation. As stated in~\cite{Jin_2018_CVPR}, this limitation is particularly problematic when a heavy blur affects the input image since the middle frame prediction becomes less reliable. Our approach is relatively robust to heavy blur as the proposed model generates frames independently from multiple decoders, therefore the error is not propagated (see \Fref{fig:errorprop}). We observed lower quantitative number in panorama scenario compared to the high speed video for both Jin \etal and our model. This is most likely because panorama GT images are relatively sharper while high speed video contains less sharp GT frames due to dynamic motion and short exposure time.

\begin{table}[!t]
\setlength{\tabcolsep}{8pt}
\begin{center}
\caption{Quantitative evaluation on GoPro blur dataset}
\label{tbl:quant2}
\vspace{-1.5mm}
  \begin{tabular}{l|lccc}
    \toprule
    & Methods & $F_i$ & $F_m$ & $F_f$  \\ 
    \midrule
    PSNR & Jin \etal  &  23.713 & 29.473 & 23.681 \\
    & Ours & \textbf{27.357} & \textbf{31.989} & \textbf{27.414} \\ \midrule
    SSIM &  Jin \etal  &  0.660 & 0.846 & 0.659 \\
    & Ours &   \textbf{0.794} & \textbf{0.885} & \textbf{0.793}\\
    \bottomrule
    \end{tabular}
\end{center}
\vspace{-6mm}
\end{table}
\vspace{-2.5mm}

\begin{figure*}[!t]
\vspace{-0.2cm}
\begin{center}
\setlength{\tabcolsep}{1pt}
\resizebox{1.0\linewidth}{!}{%
\footnotesize
\begin{tabular}{cccc|ccccc}
    Input & $F_i$ & $F_m$ & $F_f$ & Input & $F_i$ & $F_m$ & $F_f$ & \\
    \includegraphics[width=0.1\linewidth]{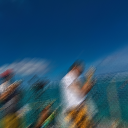} &
        \includegraphics[width=0.1\linewidth]{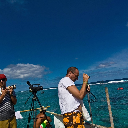} &
        \includegraphics[width=0.1\linewidth]{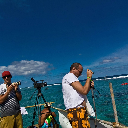} & 
        \includegraphics[width=0.1\linewidth]{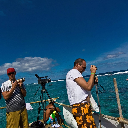} &
        \includegraphics[width=0.1\linewidth]{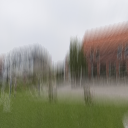} &

        \includegraphics[width=0.1\linewidth]{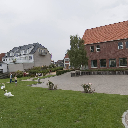} &
        \includegraphics[width=0.1\linewidth]{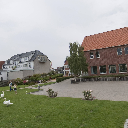} & 
        \includegraphics[width=0.1\linewidth]{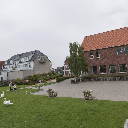} &
        \raisebox{1.7\normalbaselineskip}[0pt][0pt]{\scriptsize{{\rotatebox[origin=c]{90}{GT}}}}
        \\

        &
        \includegraphics[width=0.1\linewidth]{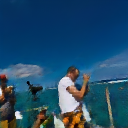} &
        \includegraphics[width=0.1\linewidth]{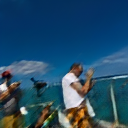} &
        \includegraphics[width=0.1\linewidth]{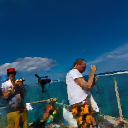} &

        &
        \includegraphics[width=0.1\linewidth]{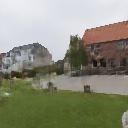} &
        \includegraphics[width=0.1\linewidth]{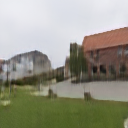} &
        \includegraphics[width=0.1\linewidth]{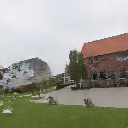} &
        \raisebox{1.7\normalbaselineskip}[0pt][0pt]{\scriptsize{\rotatebox[origin=c]{90}{Ours}} }
\end{tabular}}
\end{center}
\vspace{-3.5mm}
\caption{Rotation blurred images generated from panorama scenes. The top row is ground truth frames and the bottom row is restored frames from the blurs.}
\label{fig:qual_pano}
\vspace{-3.5mm}
\end{figure*}

\begin{figure*}[!t]
\begin{center}
\setlength{\tabcolsep}{1pt}
\resizebox{1.0\linewidth}{!}{%
\footnotesize
\begin{tabular}{cc|cc}
    Input & \hspace{-0.1cm} GT \hspace{0.9cm} Jin \etal \hspace{0.9cm} Ours&
    Input & \hspace{-0.1cm} GT \hspace{0.9cm} Jin \etal \hspace{0.9cm} Ours \\
    \includegraphics[width=0.1\linewidth]{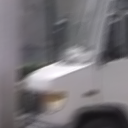} &
    \animategraphics[width=0.3\linewidth]{7}{video/1/frame-}{0}{6} &

    \includegraphics[width=0.1\linewidth]{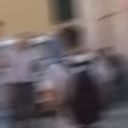} &
    \animategraphics[width=0.3\linewidth]{7}{video/2/frame-}{0}{6}

\end{tabular}}
\end{center}
\vspace{-5mm}
\caption{Heavily blurred (dynamic) inputs from the high speed videos and the restored video frames. Click on the images in \textit{Adobe Reader} to play the videos.}
\label{fig:qual_gopro1}
\vspace{-3.3mm}
\end{figure*}
\begin{figure}[!t]
\begin{center}
\setlength{\tabcolsep}{0.4pt}
\renewcommand{\arraystretch}{0.25}
\resizebox{1.0\linewidth}{!}{%
\begin{tabular}{cccc}
        \tiny
        \tiny{Input} & \tiny{$F_{i}$} & \tiny{$F_{m}$} & \tiny{$F_{f}$} \\
         \includegraphics[width=0.1\linewidth]{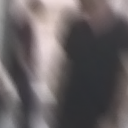} &
         \includegraphics[width=0.1\linewidth]{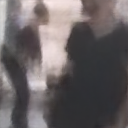} &
         \includegraphics[width=0.1\linewidth]{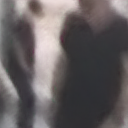} &
         \includegraphics[width=0.1\linewidth]{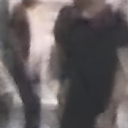} \\
         
          \includegraphics[width=0.1\linewidth]{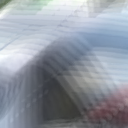} &
         \includegraphics[width=0.1\linewidth]{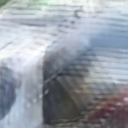} &
         \includegraphics[width=0.1\linewidth]{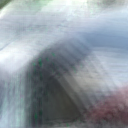} &
         \includegraphics[width=0.1\linewidth]{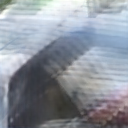} 
\end{tabular}}
\end{center}
\vspace{-4mm}
\caption{Failure cases}
\label{fig:fail_case}
\vspace{-4mm}
\end{figure}

\paragraph{Qualitative evaluation.}
The qualitative results for panoramic scenes and high speed videos show that our model can successfully restore multiple frames from a blurred input under various blur patterns (see \Fref{fig:qual_pano} and \Fref{fig:qual_gopro1}). We compare our approach and previous method \cite{Jin_2018_CVPR} on relatively heavily blurred images from the high speed video dataset. As can be seen from \Fref{fig:qual_gopro1}, our method reconstructs contents consistently across frames and restores visually sharper videos compared to \cite{Jin_2018_CVPR}. We experimentally observed that failure cases occur for temporally undersampled and severely blurred inputs (see \Fref{fig:fail_case}). The image contents of such inputs are usually destroyed, and hence, the STNs \cite{jaderberg2015spatial} and the LW layers in our network fail to learn the underlying motion from the heavily blurred inputs \ie feature decoding fails.

\subsection{Middle frame deblurring results}
\vspace{-1mm}
In addition to video restoration, we evaluate the performance of our model on image deblurring task in comparison with state-of-the-art image deblurring approaches \cite{Nah_2017_CVPR,DeblurGAN,tao2018srndeblur} on a benchmark blur dataset provided by \cite{Nah_2017_CVPR}. The dataset provides 1111 test blurred images with $1280\times 720$px resolution. We compared the middle frame prediction ($F_m$) of our pretrained 7-frame prediction model with state-of-the-art deblurring approaches and the results are summarized in \Tref{Tab:middle_comparison}. As can be inferred from \Tref{Tab:middle_comparison}, our video restoration model gives a competitive performance on image deblurring task compared to state-of-the-art deblurring approaches. The slight performance loss can be attributed to the fact our model was trained on blur dataset generated by averaging 7 frames while the benchmark dataset contains blurred images obtained by averaging more than 7 sequential frames (larger blurs).

\begin{table}[ht]
\caption{Middle frame deblurring comparison with deblurring approaches on benchmark GoPro blur dataset \cite{Nah_2017_CVPR} on PSNR metric.}
\vspace{-2mm}
\label{Tab:middle_comparison}
\centering
\resizebox{1.0\linewidth}{!}{
    \begin{tabular}{|l|l|l|l|l|}
    \hline
         \multicolumn{3}{|c|}{Single image deblurring} & \multicolumn{2}{c|}{Video restoration}  \\
         \hline
        Nah \etal  & Kupyn \etal  & Tao \etal  &  Jin \etal & Ours\\ 
        \hline
        29.08 & 28.70 & 30.26 & 26.98 & 29.84 \\
        \hline
        \end{tabular}
}
\vspace{-5mm}
\end{table}

\section{Analysis}
\begin{figure}[ht]
    \centering
    \includegraphics[width=1.0\linewidth]{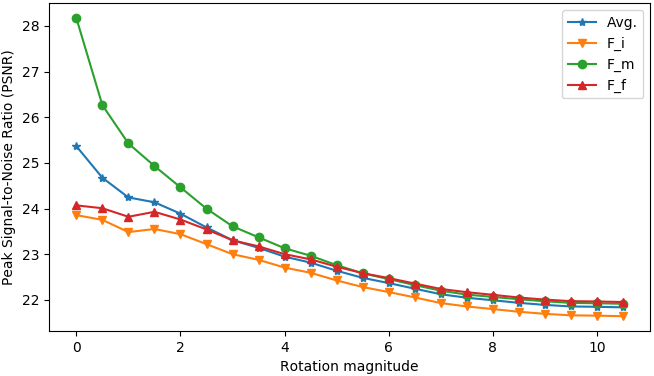}
    \caption{PSNR value vs. camera rotation magnitude for panorama test set}
    \label{fig:blur_size}
    \vspace{-2.3mm}
\end{figure}

\vspace{-0.5mm}
\paragraph{Cross-dataset evaluation.}
We report a cross-dataset  \textit{panorama}$\rightarrow$\textit{high speed video} evaluation to assess the generalization capability of our model. A model trained on the panoramic scenes is evaluated on high speed video test set (\Tref{Tab::analysis}).
Despite a performance degradation, our model trained on the panorama dataset performs on par with the competing approach \cite{Jin_2018_CVPR} trained on the high speed video dataset. The absence of dynamic motion on the panorama dataset, which is apparent in high speed videos, can be one contributing factor explaining the performance loss in addition to the domain gap \eg image contents, sharpness, blurriness.
\vspace{-3mm}
\paragraph{Size of blur.}
We analyze our model for various blur sizes by plotting the performance of the model with respect to the camera rotation magnitudes of the blurred images in the panorama test set. As can be inferred from \Fref{fig:blur_size}, the model performs better for smaller rotations and performance in general decreases for large blurs.
\vspace{-3mm}
\paragraph{Sequential error propagation.}
Previous works \cite{Jin_2018_CVPR,purohit2019bringing} are prone to error propagation as frames are reconstructed in a sequential manner starting from the middle frame. Particularly, if the deblurred middle frame is erroneous, then, the error propagates across the non-middle frames. Our work is relatively robust to sequential error propagation since all frames are predicted in a single-step without explicit middle frame dependency, hence, error does not propagate. As can be inferred from  \Fref{fig:errorprop}, for heavily blurred inputs, Jin \etal predicts erroneous middle frame and hence, the predicted non-middle frames are also erroneous. By contrast, our approach successfully recovers non-middle frames even when the middle frame prediction fails.
\vspace{-1mm}

\vspace{-3mm}
\begin{table}[ht]
\setlength{\tabcolsep}{10pt}
\caption{Quantitative results for cross-dataset evaluation}
\label{Tab::analysis}
\centering
\resizebox{0.85\linewidth}{!}{
\begin{tabular}{|lccc|}
    \hline
    \multicolumn{4}{|c|}{Panorama$\rightarrow$ high speed} \\
    \hline
       & $F_i$ & $F_m$ & $F_f$ \\ 
        \hline
        PSNR &  23.383 & 30.300 & 23.380  \\ 
        SSIM &  0.649 & 0.832& 0.651  \\
        \hline
        \end{tabular}
}
\vspace{-3.5mm}
\end{table}

\begin{figure}[!t]
\begin{center}
\setlength{\tabcolsep}{0.4pt}
\renewcommand{\arraystretch}{0.25}
\resizebox{1.0\linewidth}{!}{%
\begin{tabular}{ccccc}
        \tiny
        \tiny{Input} & \tiny{$F_{i}$} & \tiny{$F_{m}$} & \tiny{$F_{f}$}&\\ 
        \includegraphics[width=0.1\linewidth]{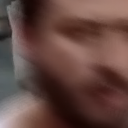} &
        \includegraphics[width=0.1\linewidth]{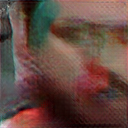}&
        \includegraphics[width=0.1\linewidth]{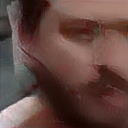}&
        \includegraphics[width=0.1\linewidth]{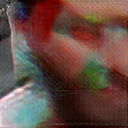}&
         \raisebox{0.8\normalbaselineskip}{\rotatebox[origin=c]{90}{\tiny{Jin \etal}}} \\
         &
        \includegraphics[width=0.1\linewidth]{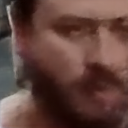}&
        \includegraphics[width=0.1\linewidth]{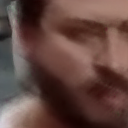}&
        \includegraphics[width=0.1\linewidth]{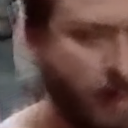}&
         \raisebox{0.8\normalbaselineskip}{\rotatebox[origin=c]{90}{\tiny{Ours}}}
\end{tabular}}
\end{center}
\vspace{-4mm}
\caption{Sequential error propagation}
\label{fig:errorprop}
\end{figure}

\section{Ablation studies}
\label{sec:ablation}
\paragraph{Network components.}
The STNs in the feature transformer network are the core part of our model for network convergence. The addition of local warping (LW) layer also significantly improves the performance of our model. The best model performance is, yet, achieved with the all network components (FTN and ITN) combined (\Tref{Tab::ablation}). The refining block improves performance by a margin of 0.43 dB on average. 
\vspace{-2mm}
\paragraph{Loss terms.}
As mentioned earlier, the multi-scale photometric loss (PML) is a sufficient constraint to make our network converge during training. We also experimentally find that a model trained with transformation consistency loss (TCL) not only converges faster with smoother behavior but also gives a better performance during testing. The penalty term (PT) gives a marginal performance improvement when predicting fewer frames as photometric loss is already a sufficient constraint (see \Tref{Tab::ablation}).
In 3 frame prediction model, the penalty term improved performance marginally around 0.25dB while in 7 frame prediction model, it improved approximately 0.6dB. Penalty term enforces the model to consider subtle differences especially when the motion is small.
\vspace{-3.5mm}

\begin{table}[!t]
\setlength{\tabcolsep}{3.5pt}
\vspace{-0.5mm}
\caption{Ablation studies on GoPro blur dataset for network components and loss terms on PSNR metric.}
\vspace{-2mm}
\label{Tab::ablation}
\centering
\resizebox{1.0\linewidth}{!}{
   \begin{tabular}{|l|l|ccc|}
 \hline
  &     & $F_i$ & $F_m$ & $F_f$  \\ 
    \hline
    & FTN [STN] &  25.86 & 31.20 & 25.78 \\
    Network& FTN [STN $\oplus$ LW] &  26.67 & 32.02 & 26.58 \\
    components& FTN [STN] $\oplus$ ITN &   26.06 & 31.78 & 26.05 \\
    & FTN [STN $\oplus$ LW] $\oplus$ ITN & 27.35 & 31.98 & 27.41 \\
    \hline
      & PML &  25.98 & 30.77 & 25.97    \\
      {Loss terms}    & PML $\oplus$ TCL & 27.08 & 31.78 & 27.12 \\
      & PML $\oplus$ TCL $\oplus$ PT & 27.35 & 31.98 & 27.41 \\
    \hline
    \end{tabular}
}
\vspace{-4.3mm}
\end{table}

\vspace{-1.5mm}
\section{Conclusion}
\vspace{-2mm}

We present a novel unified architecture that restores video frames from a single blurred image in an end-to-end manner without motion supervision. We evaluate our model on the two datasets with rotation blurs and dynamic blurs and demonstrate qualitatively and quantitatively favorable performance against the competing approach.
The cross-dataset evaluation demonstrates that our model can generalize even when the training and test set have significantly different blur patterns and domain gap. Unlike the previous approaches, our model predicts frames in a single step without middle frame dependency. It is advantageous not only because it is simple to use but also robust to heavy blurs where middle frame prediction often fails. Overall, the simplicity and flexibility of our method makes it a promising approach for future applications such as deblurring and temporal super resolution.

\paragraph{Acknowledgements.}
This work was supported by NAVER LABS Corporation [SSIM: Semantic \& scalable indoor mapping].

{\small
\bibliographystyle{ieee_fullname}
\bibliography{egbib}
}

\end{document}